# Deep Object Detection based Mitosis Analysis in Breast Cancer Histopathological Images

Anabia Sohail[2], Muhammad Ahsan Mukhtar[1], Asifullah Khan[1,2*],
Muhammad Mohsin Zafar[1], Aneela Zameer[1], Saranjam Khan[3]

[1] Pattern Recognition Lab, DCIS, PIEAS, Nilore, Islamabad 45650, Pakistan
[2] Deep Learning Lab, Center for Mathematical Sciences, PIEAS, Nilore, Islamabad 45650, Pakistan
[3] Department of Physics, Islamia College Peshawar, Pakistan
asif@pieas.edu.pk

## Abstract

Empirical evaluation of breast tissue biopsies for mitotic nuclei detection is considered an important prognostic biomarker in tumor grading and cancer progression. However, automated mitotic nuclei detection poses several challenges because of the unavailability of pixel-level annotations, different morphological configurations of mitotic nuclei, their sparse representation, and close resemblance with non-mitotic nuclei. These challenges undermine the precision of the automated detection model and thus make detection difficult in a single phase. This work proposes an end-to-end detection system for mitotic nuclei identification in breast cancer histopathological images. Deep object detection-based Mask R-CNN is adapted for mitotic nuclei detection that initially selects the candidate mitotic region with maximum recall. However, in the second phase, these candidate regions are refined by multi-object loss function to improve the precision. The performance of the proposed detection model shows improved discrimination ability (F-score of 0.86) for mitotic nuclei with significant precision (0.86) as compared to the two-stage detection models (F-score of 0.701) on TUPAC16 dataset. Promising results suggest that the deep object detection-based model has the potential to learn the characteristic features of mitotic nuclei from weakly annotated data and suggests that it can be adapted for the identification of other nuclear bodies in histopathological images.

# 1. Introduction

Breast Cancer is one of the most commonly occurring deadly diseases and is responsible for the death of more than 2 million women every year around the world [1]. Histopathological examination of tissue samples is considered a more reliable and authentic method for cancer diagnosis and characterization [2]. Empirical evaluation of mitotic activity in breast cancer patients is one of the essential prognostic biomarkers of Nottingham Histology Score for deciding the tumor grade and aggressiveness of cancer. Pathologists visually analyze tissue samples under 10 High Power Field (HPF) for assigning a score to slide. Mitotic activity is assigned a score of 1, 2, 3, respectively, for the presence of 0-11, 12-22, or more than 23 mitotic nuclei in HPF [3]. However, manual counting of mitosis is very tedious, requires expertise, and suffers from inter and intra-observer variability [4]. Mitosis detection is difficult in practice, because of its close resemblance to other cellular and nuclear bodies such as lymphocytes, apoptotic, and dense nuclei, etc. Moreover, mitotic nuclei exist under different morphological appearances as it goes through different cell differentiation stages [5].

Advancements in Machine Learning (ML) based diagnostic system and adoption of image analysis in histopathologies such as availability of imaging-based microscopy, digital slide scanners, and visualization tools have upsurge interest in development of ML-based diagnostic tools for histopathology [6]. Nowadays, deep learning techniques have surpassed conventional ML techniques and shown state-of-the-art results for medical data [7]. Different competitions on breast cancer histopathological image analysis have been organized by computer vision societies such as ICPR12, AMIDA13, ICPR14, and TUPAC16 to accelerate research in the development of automated models for mitotic nuclei characterization [6], [8]–[10]. In these competitions, deep convolutional neural networks (DCNN) based techniques appear as front-end runners and have shown exemplary performance for mitosis detection. Deep CNNs are a type of representative learning algorithms that automatically learn domain knowledge from raw pixels without putting effort in filter designing. This distinct attribute of CNN is due to the convolution operation, whereby convolution operation considers the context of features in addition to the raw representation [11]. Different CNN based classification and detection models such as VGG, ResNet, DenseNet, YOLO, faster-R-CNN, FCNN, etc. are adapted for histopathological images [12]–[17]. In this study, Mask R-CNN [18] based deep object-detection technique is proposed for mitotic nuclei analysis in histopathological images. The advantages of the proposed method are following:

(i) Reutilization of the parameters of the pre-trained network using transfer learning provide good initial set of weights for the training of deep NN for small patient dataset.

(ii) Region-based detection of mitotic nuclei helps in improving the discrimination power of the proposed detection model.

(iii) Candidate region selection and its subsequent refinement in a single end-to-end system, not only enhance the detection rate as well as improve the inference time.

## 2. Literature Review

Various ML based approaches have been proposed for the automation of mitosis detection. These approaches can be broadly categorized into three main classes: (i) patch wise classification (ii) pixel by pixel classification and (iii) region-based detection.

Previously, different image processing techniques such as color, local binary patterns (LBP), and histogram of oriented gradients (HOG) were used to learn the morphological features of the mitosis. These handcrafted features were assigned to the conventional ML models and ensemble-based approaches for the classification of mitotic nuclei.

Irshad et al. (2013) proposed a mitosis detection framework based on conventional image processing techniques [19]. The proposed method initially detected candidate regions by performing Laplacian of Gaussian followed by binary thresholding and morphological operations. Whereas, the color space of histopathological images was exploited to capture the statistical and morphological features to classify the mitotic regions. The proposed technique achieved an F-score of 0.72 on the MITOS12 dataset [19]. Irshad et al. extended their study and showed that the use of morphological features along with multispectral statistical features extracted from selected spectral bands improves the mitosis detection task with an F-score of 0.76 on MITOS12 multispectral dataset [20], [21]. Tek et al. [22] considered the shape and color based features and developed an ensemble of single and cascade of AdaBoosts for the classification of mitosis. The proposed technique was tested in ICPR12 competition on which it achieved 0.397 and 0.58 F-score for single and cascade of ensembles, respectively. In the study by Huang et al. [23], mitosis were discriminated by developing a novel algorithm XICA. The proposed approach was based on independent component analysis that finds the class discriminating components. Taskh et al. initially preprocessed histopathological images using morphological and 2-D anisotropic diffusion process [24]. Whereas in the second phase, SVM based classification is performed by extracting texture and color-based features using LBP and statistical approaches. However, the main concern associated with conventional ML techniques is that the performance of the detection module is dependent upon feature descriptors. Therefore, in handcrafted features, all efforts are put in the designing of filters that require expertise and domain knowledge. As mitotic objects vary in size, shapes, and texture, thus conventional algorithms show low detection accuracy.

CNN based approaches have shown promising results for vision-based problems. Different researchers exploited CNN for mitosis detection, and consequently, based on the exemplary performance shown by CNN based models, they have been placed at the top position in mitosis detection challenges. Contrary to conventional ML techniques, Ciresan et al. suggested the use of max-pooling in CNN for the classification of mitosis nuclei and stood top rank position in ICPR 2012 challenge. Similarly, Veta et al. (2016) developed six layers deep CNN for mitosis counting and validated the model on the AMIDA13 challenge dataset [25]. The proposed model was robust against color variation of histopathological images and suggested that small size mitotic nuclei are usually responsible for the poor performance of the model. Chen et al. [26] proposed cascaded CNN to tackle the mitosis detection problem. The proposed model classifies the mitosis in two stages. In the first part, candidate nuclei consisting of both actual mitosis and hard negative examples were identified using a coarse retrieval model. Whereas, in the second phase, initially generated candidate nuclei were mined by developing a new algorithm to extract the hard examples. Their proposed technique was evaluated on the Mitos12 and 14 datasets.

Contrary to discrimination of image patches as mitotic or non-mitotic, Ciresan et al. performed pixel by pixel classification using deep CNN [27]. The proposed mitosis identification technique achieved an F-score of 0.78 and 0.61 on the ICPR12 and MICCAI13 contest dataset, respectively. However, pixel-wise image classification using deep CNN is computationally expensive and time-taking. Therefore, such techniques are infeasible for the development of end-to-end system for whole-slide images that consists of several HPFs. A similar work [28] proposed a regression-based method for counting cells in microscopic images. They developed a fully convolutional regression network that generates a density map for each input image, thus eliminating a prior detection and segmentation phase. During training, cells are represented by density maps that show dot annotations in the form of Gaussian. The model is trained by minimizing the mean squared error between ground truth and predicted a density map.

Many researchers improved the precision of mitosis detection tasks by dividing the learning of the model into two different phases. Wahab et al. improved the performance of mitosis detection problem by proposing a two-phase learning strategy [29]. In the first phase, they extracted candidate mitotic nuclei consisting of both true mitotic and hard negative non-mitotic patches using pixel by pixel classification approach. Whereas, in the second phase, they developed a deep CNN based classification approach that further refines the output of the first phase. Likewise, Paeng et al. also tackled the mitosis detection problem by dividing the learning into two stages and stood first place in TUPAC 2016 challenge [30]. In the first step, they extracted positive examples using ground truth and randomly selected negative examples to train the ResNet. This step was used to generate a new dataset consisting of both original examples as well as hard negative examples produced from first step and their color variants. This augmented dataset was used to train the ResNet based CNN from scratch. In the final step, the trained model was converted into FCN to make it applicable for any size of an image.

Li et al. suggested the use of a deep detection-based framework for mitosis candidate identification [31]. They exploited the idea of weak labels and developed a deep residual network that is sensitive to hard negative examples. Their algorithm achieved acceptable results on two different datasets [31]. Li et al. extended their approach and proposed a concentric loss-based approach for mitotic nuclei classification and achieved significant results on challenging TUPAC16 dataset [32].

## 3. Materials and Methodology

Mitosis detection is a challenging task due to heterogeneous morphological appearances of nuclei and their close resemblance with non-mitotic nuclei (shown in Figure 1). This work suggests the use of transfer learning and deep object-detection technique for mitotic nuclei classification and analysis. The overall workflow of the proposed detection model is shown in Figure 2.

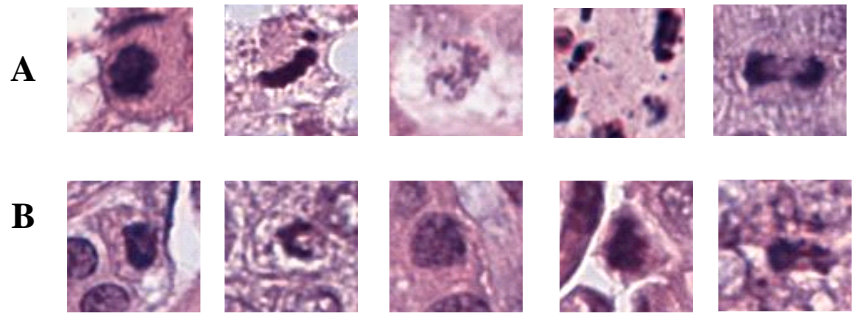

**Figure 1:** Mitosis (panel A) and non-mitosis images (panel B), showing inter-class similarity and intra-class heterogeneity.

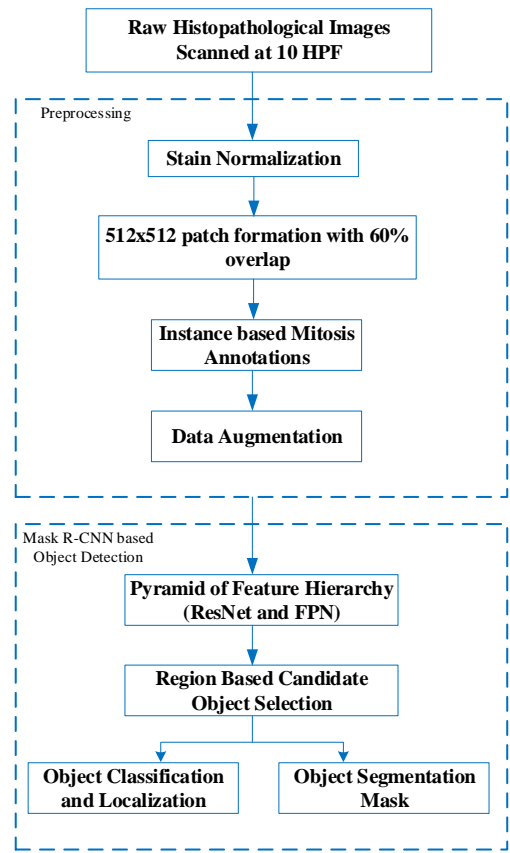

**Figure 2:** Overall workflow of the proposed detection model.

## 3.1. Dataset

The mitosis dataset from the TUPAC16 challenge was utilized for the mitosis detection task. TUPAC16 dataset is comprised of 73 patients with a varying number of patches per patient. Out of 73, slides from 23 patients are scanned with the Aperio ScanScope XT slide scanner. The rest of the 50 slides of the TUPAC16 dataset are scanned with a Leica SCN400 slide scanner. Annotations of mitosis samples were provided by at least two pathologists [10]. The training dataset was augmented by combining patient samples from two previous competitions dataset, MITOS12 and MITOS14, to improve the learning of the training phase, shown in Figure 3. Dataset division for train, validation, and test are mentioned in Table 1. In all these datasets, slides for each patient are scanned at 40x magnification and represents the tissue that corresponds to 10 HPFs. However, the resolution of samples varies, such as TUPAC16 covers approximately an area of 2mm$^2$, whereas for Aperio Scanscope samples, 1pixel = 0.2456µm and for Hamamatsu Nanozoomer, 1pixel = 0.2273µm resolution. Dataset details are mentioned in Table 2. Different patient samples from different scanners and Labs are shown in Figure 4.

**Table 1:** Distribution of patients for train, validation and test dataset.

| Dataset | Patient Number |
|---|---|
| Train | TUPAC16 patients: 01, 02, 03, 05, 07, 08, 10, 11, 12, 13, 14, 15, 16, 17, 18, 19, 20, 22, 23, 24, 25, 28, 33, 34, 35, 37, 38, 40, 42, 44, 47, 49, 51, 52, 54, 59, 61, 64, 68, 69, 70, 72<br>All patients from training dataset of MITOS12 and MITOS14 |
| Validation | 04, 06, 09, 21, 26, 29, 31, 39, 46, 48, 56, 65, 67, 73 |
| Test | 27, 30, 32, 36, 41, 43, 45, 50, 53, 55, 57, 58, 60, 62, 63, 66, 71 |

**Table 2:** Dataset details. Tissue slides were scanned at 40x magnification under 10 HPF.

| Dataset | Scanner | Resolution | Spatial Dimension | Patients |
|---|---|---|---|---|
| Tupac-16 | Aperio ScanScope | 0.25 $\mu m$ /pixel | 2000x2000 | 23 |
| | Leica SCN400 | 0.25 $\mu m$ /pixel | 5657x5657 | 50 |
| Mitos12 | Leica SCN400 | 0.2456 $\mu m$ /pixel | 2084x2084 | 5 |
| | Hamamatsu Nanozoomer | 0.2275 $\mu m$ /pixel | 2252x2250 | 5 |
| Mitos14 | Leica SCN400 | 0.2456 $\mu m$ /pixel | 1539x1376 | 11 |
| | Hamamatsu Nanozoomer | 0.2275 $\mu m$ /pixel | 1663x1485 | 11 |

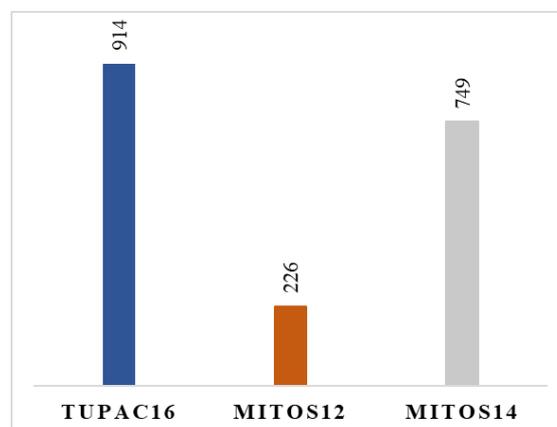

**Figure 3:** Mitosis frequency in the training dataset.

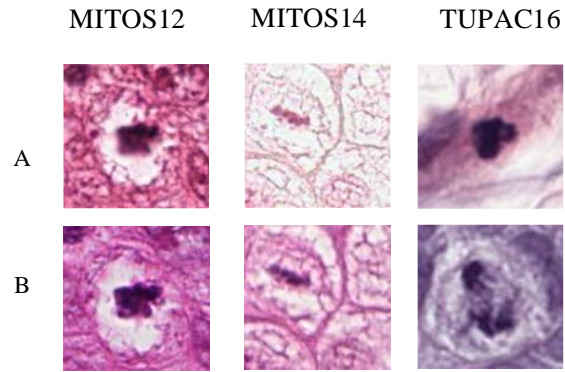

**Figure 4.** Mitotic nuclei from three different datasets: MITOS12, MITOS14, and TUPAC16 are shown. Images from Leica and Hamamatsu Scanner are shown for the MITOS12 and Mitos14 datset, respectively in panel A and B. Whereas images for TUPAC16 dataset are shown from Aperio ScanScope XT and Leica Scanner, respectively in panel A and B.

### 3.2. Preprocessing and Normalization of Images

The patient's samples (histopathological images) are usually collected from different labs that result in variation in the stain color of samples (shown in Figure 5). Similarly, the use of different scanners adds noise to images (shown in Figure 4). In this regard, stain normalization is performed using [33] to reduce the contrasting variation between the slides, as shown in Figure 5. In order to reduce the effect of noise, pixel-wise mean normalization is performed. Due to memory constraints and limited processing power, images were cropped into 512x512 size. Moreover, to avoid cutting off the mitotic nuclei during patch extraction, a sliding window approach with an overlap of 60% was used. The training dataset was augmented by applying a horizontal flip.

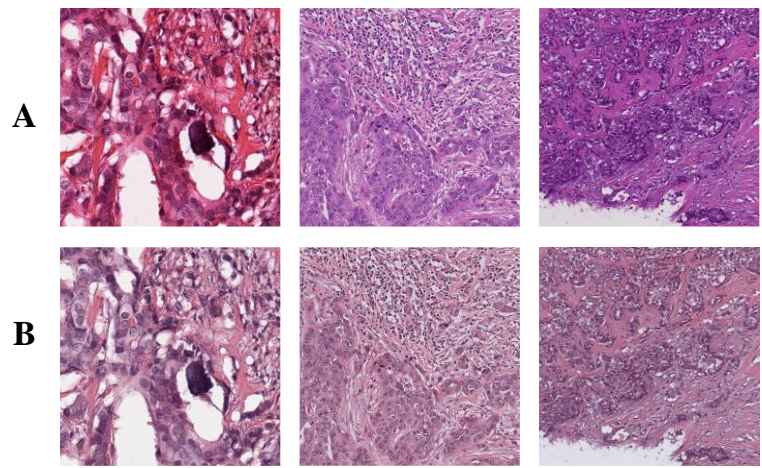

**Figure 5**: Histopathological samples without stain normalization are shown in panel A whereas panel B shows images after applying stain normalization.

## 3.3. Mask Formation

Annotations of Mitos14 and TUPAC16 dataset are provided as weak labels in the form of centroid position of the mitosis, shown in Figure 6. Weakly labeled data is unable to give the precise information about the characteristic features of mitosis that is required for the training of the detection model. Mitotic nuclei have no well-defined shape. However, during different phases of cell division, mitosis, to some extent, morphologically resembles ellipse or circle (Figure 1). Labels for mitotic nuclei are developed from the weak annotations by adapting the idea proposed by [32] and represented each mitosis in the form of a circle or ellipse. Circular annotation is drawn randomly with a radius of 10-16 pixels. Similarly, elliptical shape is drawn by randomly drawing pixels in the range of 5-13 pixels in the direction of 60° and 90° angle. Some examples of generated masks are shown in Figure 7.

## 3.4. Two Stage Detection Framework

Mitosis detection is a class imbalance problem, which is overwhelmed by a large number of non-dividing nuclei. This imbalance makes classifier bias towards the majority class. In this regard, the two-phase detection framework is proposed in the literature whereby to improve the performance of the classifier on the imbalanced dataset [30], [31]. Therefore, the two-phase detection framework proposed by Wahab et al. is used in this study to mitigate the class imbalance effect (shown in Figure 8) [34]. Following their approach, in the first step, FCN has used for the detection of candidate mitotic nuclei. The output of the first phase is used as the input of the second phase, and it is refined to remove false positive by assigning it to a deep CNN classifier. This approach reduced class imbalance by producing 1:16 mitosis vs. non-mitosis examples. For the detection phase, ImageNet pre-trained VGG-16 network is used that consisted of 13 convolutional and 3 fully connected layers. This architecture was fine-tuned using breast cancer histopathological images of $512 \times 512$ patch size. In the second phase, mitotic nuclei are discriminated from hard examples by using ResNet101.

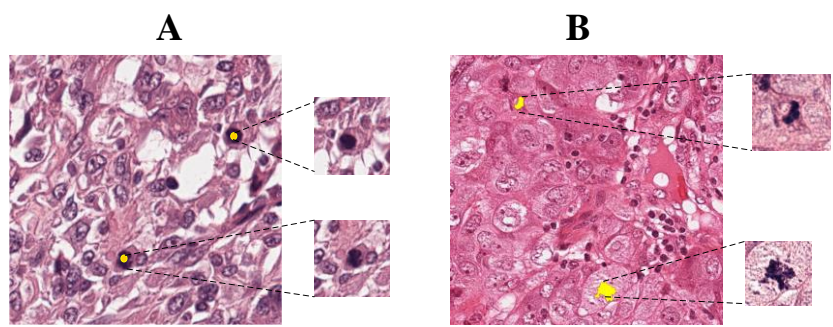

**Figure 6:** Weak labels (panel A) vs pixel level labels (panel B)

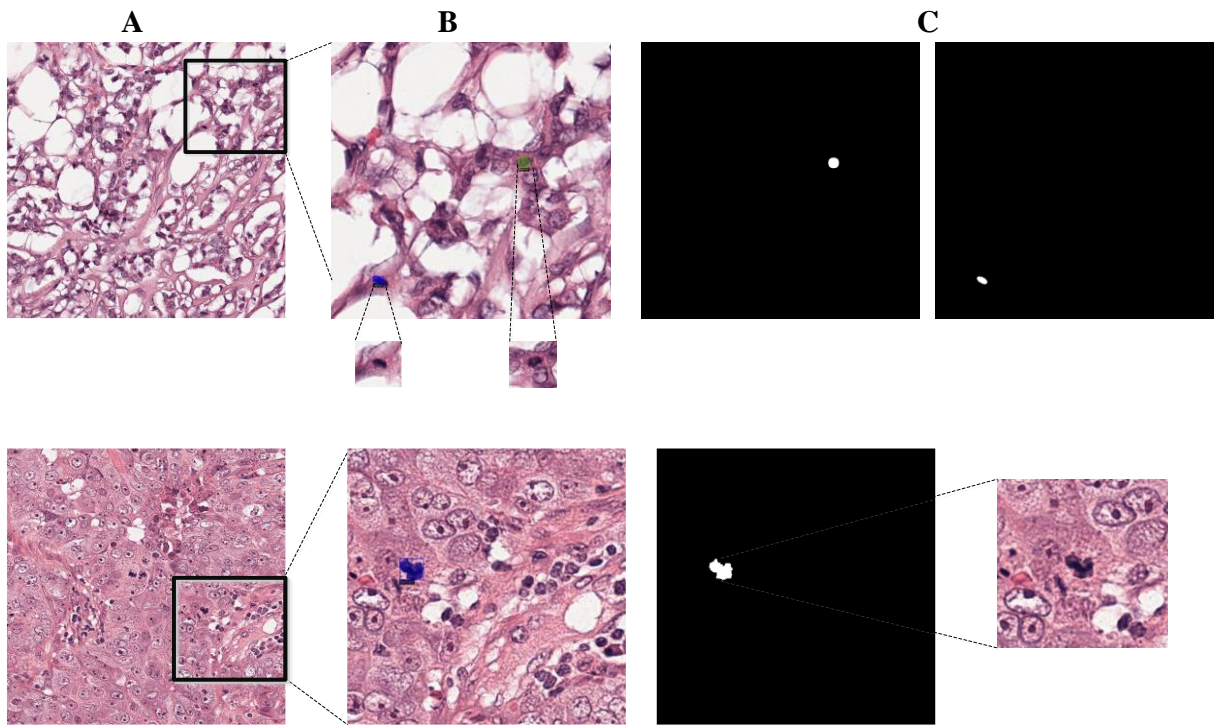

**Figure 7:** Original image (panel A), label and bounding box for mitosis (panel B) and their corresponding masks (panel C) are shown.

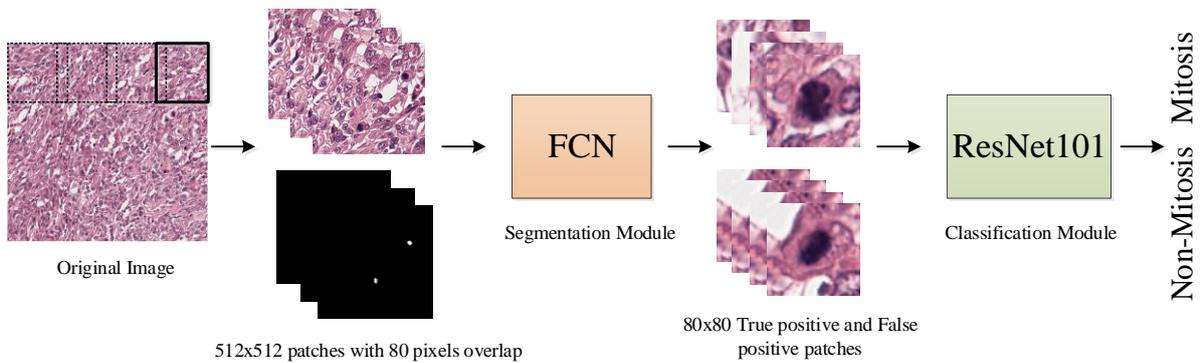

**Figure 8:** Two-stage detection framework.

### 3.5. Deep Object Detection based Analysis of Mitotic Nuclei

Mitosis classification is inherently an object detection problem. Therefore, mitotic regions are labeled as the object of interest, whereas all the remaining objects and surrounding regions are considered as background. Mask R-CNN is used simultaneously for the detection and segmentation of mitotic nuclei in histopathological images. Mask R-CNN is a region-based segmentation framework that uses multi-objective loss function to perform object classification, bounding box regression, and segmentation simultaneously. The loss function of Mask R-CNN is mathematically expressed in equation (1).

$$E = E_{cls} + E_{reg} + E_{mask} \qquad (1)$$

The learning framework of Mask R-CNN is divided into (i) Backbone Architecture, (ii) Region Proposal Network (RPN), (iii) Region of Interest (ROI) alignment, (iv) Detection, and Segmentation. The block diagram of the Mask R-CNN architecture is shown in Figure 9. In the first stage, enriched feature hierarchies, in combination with high-level semantic information, are scanned to select the region proposals. The second phase classifies the proposal and generates the bounding box and mask.

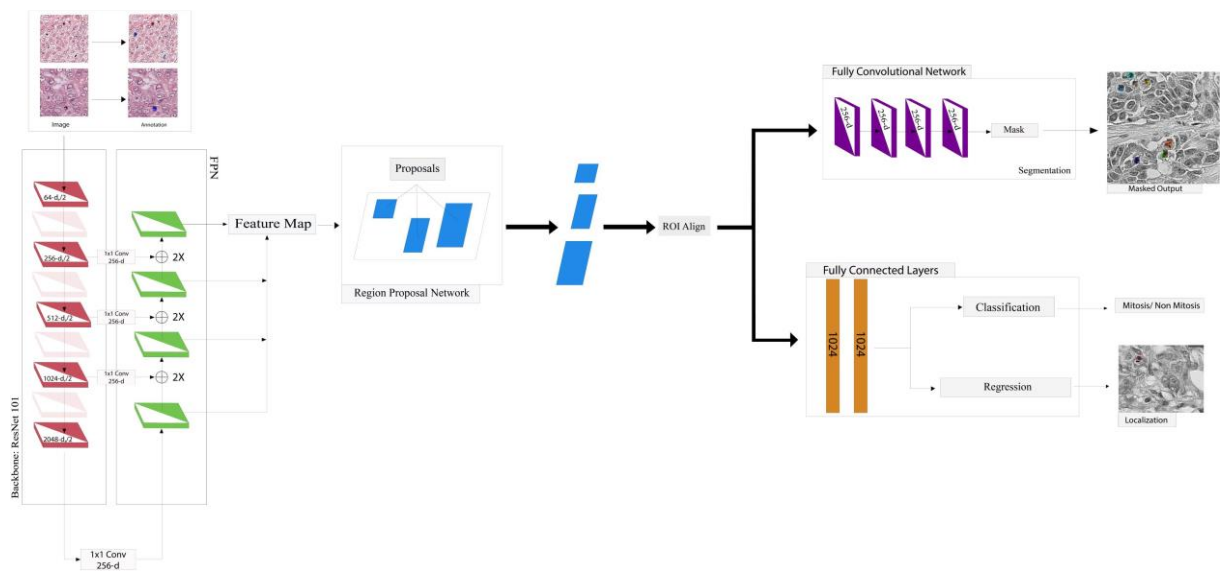

**Figure 9:** Block Diagram of the proposed detection Model.

### 3.5.1. Backbone Network

For mitosis detection, ResNet101 is used as a backbone architecture in Mask R-CNN to learn the semantic information of the object by performing feature propagation in a bottom-up fashion [35]. However, this forward flow loses the spatial information, which is required for object detection. Moreover, mitosis vary in size and shape and usually small in size. Therefore, feature pyramid network (FPN) is used in combination with ResNet101 to reinstate the spatial relation of the object with semantic information [36]. FPN rescales the feature map by interpolating the feature-map in a top-down manner and adding preliminary information by drawing residual connections from the ResNet101 backbone. FPN outputs the multi-scale and multi-level information that enables the RPN to tackle objects of variable sizes.

### 3.5.2. RPN

RPN is a small network that slides across the feature-maps to select the sub-regions of an image that are most likely to contain object responses. As mitosis are irregular in shape and vary in size therefore object response is evaluated on four different scales of {32 × 32, 64 × 64, 128 × 128, 256 × 256} and three different aspect ratios of {1:2,1:1,2:1}. In this regard, twelve different anchors are used to extract the sub-regions. Region proposals are selected by computing Intersection over Union (IoU) between the anchored region and ground-truth whereby only those sub-regions are passed towards the next phase that overlaps more than a given threshold, which in our case is 70%.

### 3.5.3. ROI Alignment, Detection and Segmentation

The output-response of RPN is variable size, rectangular sub-regions of feature-maps that are known as region proposal. These region proposals are need to be realigned before assigning to detection and segmentation branch. ROI Alignment pools out all the region proposals and feature-maps to a similar size by using bi-linear interpolation to include the contribution of all the regions, thereby solving the problem of information loss. The output from ROI Alignment is passed to fully connected layers for bounding box localization and classification. The error function for classification and bounding box localization is expressed in equation (2). Whereas, simultaneously, selected mitotic regions are segmented by performing pixel-level classification using FCN.

$$E_{cls+reg} = \frac{1}{N_{cls}} \sum_k E_{cls}(C_k, C_k^*) + \lambda \frac{1}{N_{reg}} \sum_k C_k^* E_{reg}(B_k, B_k^*) \qquad (2)$$

$$E_{cls} = -\log P(C_k^* | C_k) \qquad (3)$$

$$E_{reg} = smooth L_1 (B_k - B_k^*) \qquad (4)$$

$$smooth L_1 (B_k - B_k^*) = \begin{cases} 0.5(B_k - B_k^*) & if \ |B_k - B_k^*| < 1 \\ |B_k - B_k^*| - 0.5 & otherwise, \end{cases} \qquad (5)$$

$$E_{mask}(C^*, C) = \frac{1}{W \times H} \sum_{m=1}^{M} \sum_{n=1}^{N} \left[ C_{m,n}^* . \log C_{m,n} + (1 - C_{m,n}^*) . \log(1 - C_{m,n}) \right] \qquad (6)$$

In equation (2), $E_{cls}$ shows the classification error for region proposal that is computed using log loss (shown in equation (3)). $C_i$ is the probability of $k^{th}$ anchor being predicted as mitosis, whereas $C_i^*$ represents the ground-truth label for $k^{th}$ anchor. Similarly, $E_{reg}$ uses smooth L1 loss equation (4 & 5) to minimize the difference between predicted $B_i$ and actual $B_i^*$ bounding box coordinate positions for $k^{th}$ anchor. Whereas, $N_{cls}$ and $N_{reg}$ are used as normalizing factor that represents the size of a mini-batch, and number of anchor locations, respectively. Trade-off between the two losses is made by using $\lambda$ as hyper-parameter. On the other hand, segmentation mask for mitotic nuclei are generated by evaluating pixel-level classification using binary cross entropy loss function as shown in equation (6). Where WxH represents the spatial dimension of input image and $C_{m,n}^*$ and $C_{m,n}$ shows ground-truth and predicted probability at *m, n* coordinates of the image.

## 3.6. Transfer Learning

Deep NNs have large number of parameters and require huge amount of data for training. However, histopathological samples of patient are usually sparse in number, therefore training of deep NN from scratch on histopathological images usually results in suboptimal performance and poor convergence. Transfer learning is one of the effective ways of improving the learning of deep NNs by transferring the knowledge of the pre-trained network known as source domain to the target domain [37]. Moreover, a distinct characteristic of images is that preliminary features such as curves, semi-circles, circles, lines, edges, and color are shared among diverse categories of images [38]. This relatedness allows reusing the parameters of a network trained on natural images to the medical image dataset.

In this study, idea of transfer learning is exploited to get good initial set of weights from the pre-trained network to improve the convergence of deep object detection model on small training set. In this work, source domain $D(S)$, which is consisted of natural images of objects is defined by feature space $X_s$ and its marginal distribution $P(X_s)$ (mathematically expressed in equation (7)). Whereas the target domain $D(T_m)$ is defined by mitotic and non-mitotic images $X_m$ and their marginal probability $P(X_m)$, which is represented by equation (8). This work hypothesizes that the transfer of parameters $\theta_S$ from a network $M_s = f_s(\theta_s, X_s)$ pre-trained on the source domain to the target domain network $M_T = f_T(\theta_T, X_m)$ can help to improve the learning of deep NN for small datasets. Therefore, a backbone architecture pre-trained on ImageNet and detection network pre-trained on the COCO dataset is fine-tuned on histopathological images of breast cancer patients. This fine-tuning of a pre-trained network on histopathological images using multi-objective loss function adapts the parameters of the network according to the target domain.

$$D(S) = \{X_s, P(X_s)\} \tag{7}$$

$$D(T) = \{X_m, P(X_m)\} \tag{8}$$

$$M_s = f_s(\theta_s, X_s) \tag{9}$$

$$M_T = f_T(\theta_T, X_m) \tag{10}$$

## 3.7. Performance Metrics

The performance of the proposed approach is evaluated based on the F-score. The F-score is the harmonic mean of precision and recall (equation (13)), which handles the class imbalance problem. Mitosis in the last phase of cell division (telophase) are dividing as two, and thereby two nuclei are counted as single. Therefore, according to the criterion of the automated mitosis detection models, any prediction that is made within 30 pixels of ground truth centroid is consider as true mitosis and counted once.

As we are interested in the detection model that accurately detects the mitosis with a minimum number of false positives; therefore, precision and recall are also considered along with F-score. The recall is used to determine the detection capacity of the proposed approach, whereas precision is used to measure how close model predictions to the actual class label are. Performance matrices are mathematically expressed in equation 11-13.

$$\text{Recall} = \frac{TP}{TP + FN} \tag{11}$$

$$\text{Precision} = \frac{TP}{TP + FP} \tag{12}$$

$$F - Score = 2 \cdot \frac{\text{Recall} \times \text{Precison}}{\text{Recall} + \text{Precison}} \tag{13}$$

In the above equation, TP and FP suggest the predicted mitosis, where TP are the mitosis that actually belong to mitosis class, and FP are examples that are incorrectly classified as mitosis. Whereas FNs are positive examples that are incorrectly identified as negative.

## 3.8. Parameter Settings

ResNet101 was trained using cross-entropy loss function by setting batch size equal to 12, assigning a learning rate of 1e-4, and weight decay of 1e-2 for 100 epochs. During training, early stopping criterion is used. Network training is stopped when no significant improvement on the validation set is noted, and all training examples are completely learned by the network.

Mask R-CNN was trained for 30,000 epochs using batch size 2, and the learning rate equal to 0.0025. Region proposals were selected based on a threshold value greater than 0.7 for IoU of ground truth and anchored region. The number of region proposals was set to be 78 for mitosis and non-mitosis examples for assigning to ROI head for classification. ResNet101 and Mask R-CNN were optimized using Adam and SGD optimizers, respectively.

### 3.9. Implementation Details

All the experiments were performed in PyTorch framework and run on NVIDIA GeForce GTX 1070 having 8GB and Tesla-K80 on google-Collaboratory. Whereas stain normalization of images was performed by running script in MATLAB2016A on a desktop computer, having 16 GB RAM, Intel(R) Core(TM) i7-33770, 64-bit operating system, CPU@3.4 GHz.

## 4. Results

Automated detection of mitotic nuclei is a challenging task because of sparse representation and close resemblance of dividing nuclei with non-mitotic nuclei. In this work, Mask R-CNN based, an end-to-end system is proposed for mitosis detection. Since a limited number of patient samples are available, that may cause over-fitting of a model without learning of intrinsic representation of mitosis. Therefore, parameters of deep NNs that were pre-trained on the large dataset of natural images and diverse categories of objects are re-utilized for mitosis specific feature learning. Performance of Mask R-CNN is evaluated on object level as a focus of research is on determining the exitance or absence of mitosis in a specific region rather than its exact shape. Results of Mask R-CNN are shown in Figure 10. Detected mitosis are shown by a rectangular box, whereas the number on the rectangular frame shows the confidence score for the predicted mitosis class. Similarly, segmentation of mitosis is shown in the form of a binary mask that outlines the mitosis.

### 4.1. Comparison with two-Stage Detection Framework

Performance of Mask R-CNN is compared with the two-stage detection model to evaluate its potential in candidate mitotic region selection and detection. For a fair comparison, ResNet101, which was a backbone architecture of Mask R-CNN, is used as a feature extractor in the two-stage detection model. Mask R-CNN shows better performance in terms of F-score, precision, and Recall, as shown in Table 3. Recall clearly shows an improved detection rate.

### 4.2. Comparison with the Existing Techniques

The proposed mitosis detection framework is compared with the existing techniques on the TUPAC16 dataset to evaluate its performance (shown in Table 4). Performance comparison shows the promising performance of the proposed technique in terms of F-score as compared to existing approaches (shown in Figure 11). Performance analysis suggests that the use of multi-objective loss function by Mask R-CNN not only helps in the selection of appropriate mitosis and non-mitosis region proposals but also have good discrimination power.

## 5. Conclusion

Mitosis count is one of the important prognostic markers of the Nottingham Histology grading scheme for cancer. Automation of mitosis detection can facilitate the pathologists in resolving the intra and inter-observer conflict. In this work, we adapt Mask R-CNN for mitosis detection and demarcation of the mitotic region boundary. Parameters of the pre-trained model are exploited using transfer learning for the efficient training of deep object detection model. Whereas, Mask R-CNN is adapted to mitosis data by fine-tuning it on breast cancer histopathological images. Mitotic nuclei are characterized by different morphological appearances in four different stages of cell division. Moreover, pathologists as centroid annotation of ground truth usually weakly label mitotic nuclei. Thereby, these challenges affect the generalization of the classifier and make automation of the manual process a challenging task.

The proposed detection scheme shows promising results with an F-score of 0.86 for mitosis detection on the TUPAC16 dataset. Moreover, a comparison of the Mask R-CNN with existing techniques shows adequate learning capacity for weakly labeled datasets. It suggests that it has the potential to learn characteristic features of mitosis regardless of the unavailability of pixel-level annotation of mitosis morphology. The performance of the proposed scheme suggests that in the future, Mask R-CNN can be exploited for the detection and segmentation of other weakly annotated nuclear bodies in histopathological images.

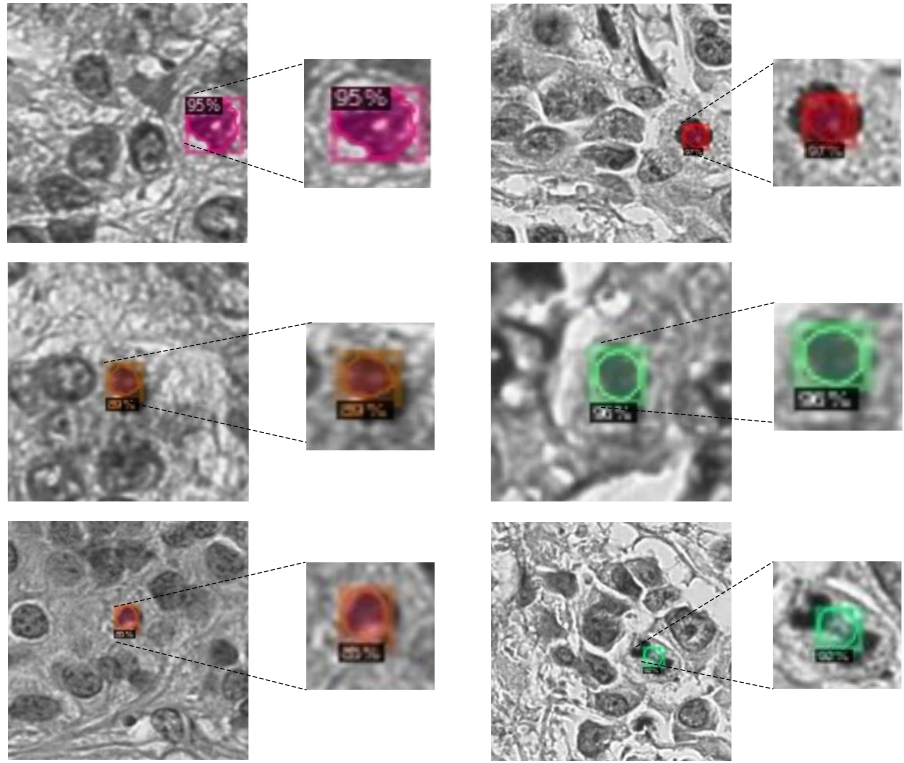

**Figure 10:** Results of Mask R-CNN. Rectangular box shows the detected mitosis whereas number on the rectangular frame represents the confidence score for the predicted mitosis class. Segmentation of mitosis is shown in the form of a binary mask that outlines the mitosis.

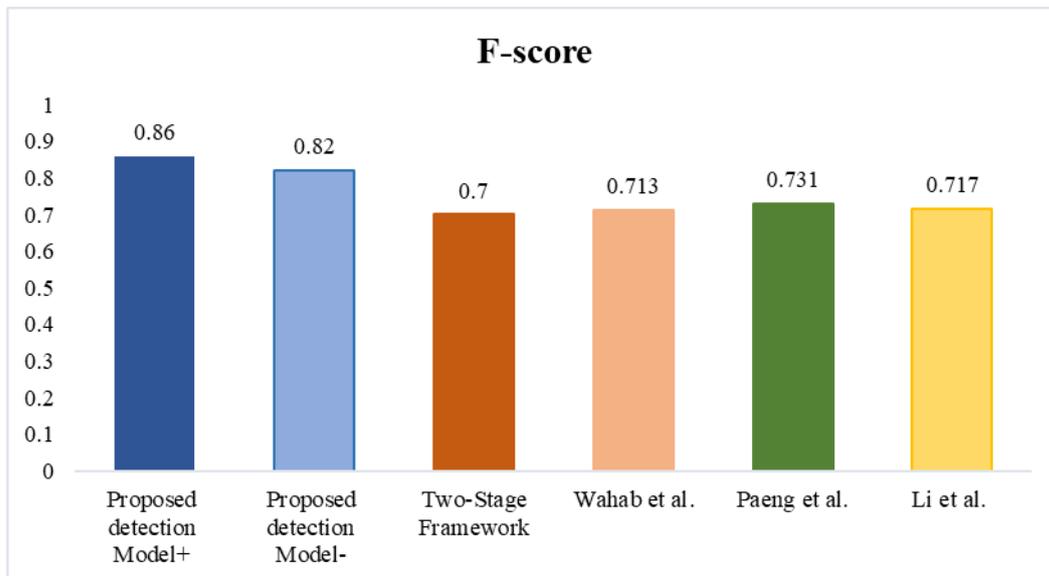

**Figure 11:** Comparison of the proposed technique with existing techniques in terms of F-score.

Table 3: Comparison of proposed Mitosis Detection model with 2 stage detection model

| Detection Model | F-Score | Precision | Recall |
|---|---|---|---|
| Proposed detection Model [+] | 0.86 | 0.86 | 0.86 |
| Proposed detection Model [-] | 0.82 | 0.798 | 0.86 |
| 2 Stage ResNet101 | 0.708 | 0.76 | 0.66 |

Table 4: Comparison of proposed Mitosis Detection model with already existing techniques

| Detection Model | F-Score | Precision | Recall |
|---|---|---|---|
| Proposed detection Model [+] | 0.86 | 0.86 | 0.86 |
| Wahab et al.[*] [29] | 0.713 | 0.77 | 0.66 |
| Li et al.[*] [32] | 0.717 | - | - |
| Paeng et al.[*] [30] | 0.731 | - | - |

## Acknowledgements

The authors thank Higher Education Commission of Pakistan (HEC) for granting funds under HEC indigenous scholarship program and Pattern Recognition lab at DCIS, PIEAS, for providing computational facilities.

## References

[1] F. Bray, J. Ferlay, I. Soerjomataram, R. L. Siegel, L. A. Torre, and A. Jemal, "Global cancer statistics 2018: GLOBOCAN estimates of incidence and mortality worldwide for 36 cancers in 185 countries.," *CA. Cancer J. Clin.*, vol. 68, no. 6, pp. 394–424, 2018.

[2] G. Stålhammar, K. Bartuma, C. All-Eriksson, and S. Seregard, "Cancer Pathology," in *Clinical Ophthalmic Oncology*, Springer, 2019, pp. 19–32.

[3] C. W. ELSTON and I. O. ELLIS, "pathological prognostic factors in breast cancer. I. The value of histological grade in breast cancer: experience from a large study with long-term follow-up," *Histopathology*, 1991.

[4] M. Veta, P. J. Van Diest, M. Jiwa, S. Al-Janabi, and J. P. W. Pluim, "Mitosis counting in breast cancer: Object-level interobserver agreement and comparison to an automatic method," *PLoS One*, 2016.

[5]     R. E. Yancey, "Multi-stream Faster RCNN for Mitosis Counting in Breast Cancer Images," pp. 1–16, Feb. 2020.

[6]     M. Veta *et al.*, "Assessment of algorithms for mitosis detection in breast cancer histopathology images," *Med. Image Anal.*, 2015.

[7]     Y. LeCun, Y. Bengio, and G. Hinton, "Deep learning," *Nature*, vol. 521, no. 7553, pp. 436–444, May 2015.

[8]     L. Roux *et al.*, "Mitosis detection in breast cancer histological images An ICPR 2012 contest.," *J. Pathol. Inform.*, vol. 4, no. 1, p. 8, 2013.

[9]     L. Roux, "Mitosis atypia 14 grand challenge." 2014.

[10]    "Dataset | Tumor Proliferation Assessment Challenge 2016." .

[11]    A. Khan, A. Sohail, U. Zahoora, and A. S. Qureshi, "A Survey of the Recent Architectures of Deep Convolutional Neural Networks," *arXiv Prepr. arXiv1901.06032*, pp. 1–60, Jan. 2019.

[12]    E. Zerhouni, D. Lanyi, M. Viana, and M. Gabrani, "Wide residual networks for mitosis detection," in *Proceedings - International Symposium on Biomedical Imaging*, 2017.

[13]    G. Huang, Z. Liu, L. Van Der Maaten, and K. Q. Weinberger, "Densely connected convolutional networks," *Proc. - 30th IEEE Conf. Comput. Vis. Pattern Recognition, CVPR 2017*, vol. 2017-Janua, pp. 2261–2269, 2017.

[14]    E. Shelhamer, J. Long, and T. Darrell, "Fully Convolutional Networks for Semantic Segmentation," *IEEE Trans. Pattern Anal. Mach. Intell.*, vol. 39, no. 4, pp. 640–651, 2017.

[15]    R. Girshick, "Fast R-CNN," in *Proceedings of the IEEE International Conference on Computer Vision*, 2015.

[16]    K. Simonyan and A. Zisserman, "VERY DEEP CONVOLUTIONAL NETWORKS FOR LARGE-SCALE IMAGE RECOGNITION," *ICLR*, vol. 75, no. 6, pp. 398–406, 2015.

[17]    C. Szegedy, S. Ioffe, and V. Vanhoucke, "Inception-v4, Inception-ResNet and the Impact of Residual Connections on Learning," *arXiv Prepr. arXiv1602.07261v2*, vol. 131, no. 2, pp. 262–263, 2016.

[18]    K. He, G. Gkioxari, P. Dollar, and R. Girshick, "Mask R-CNN," in *Proceedings of the IEEE International Conference on Computer Vision*, 2017.

[19]    H. Irshad, "Automated mitosis detection in histopathology using morphological and multi-channel statistics features," *J. Pathol. Inform.*, vol. 4, 2013.

[20]    H. Irshad, A. Gouaillard, L. Roux, and D. Racoceanu, "Spectral band selection for mitosis detection in histopathology," in *2014 IEEE 11th International Symposium on Biomedical Imaging (ISBI)*, 2014, pp. 1279–1282.

[21]    H. Irshad, A. Gouaillard, L. Roux, and D. Racoceanu, "Multispectral band selection and spatial characterization: Application to mitosis detection in breast cancer

histopathology," *Comput. Med. Imaging Graph.*, vol. 38, no. 5, pp. 390–402, 2014.

[22] F. B. Tek, "Mitosis detection using generic features and an ensemble of cascade adaboosts," *J. Pathol. Inform.*, vol. 4, 2013.

[23] C.-H. Huang and H.-K. Lee, "Automated mitosis detection based on exclusive independent component analysis," in *Proceedings of the 21st International Conference on Pattern Recognition (ICPR2012)*, 2012, pp. 1856–1859.

[24] A. Tashk, M. S. Helfroush, H. Danyali, and M. Akbarzadeh, "An automatic mitosis detection method for breast cancer histopathology slide images based on objective and pixel-wise textural features classification," in *The 5th conference on information and knowledge technology*, 2013, pp. 406–410.

[25] M. Veta, P. J. Van Diest, M. Jiwa, S. Al-janabi, and J. P. W. Pluim, "Mitosis Counting in Breast Cancer : Object- Level Interobserver Agreement and Comparison to an Automatic Method," pp. 1–13, 2016.

[26] H. Chen, Q. Dou, X. Wang, J. Qin, and P. A. Heng, "Mitosis detection in breast cancer histology images via deep cascaded networks," in *Thirtieth AAAI Conference on Artificial Intelligence*, 2016.

[27] D. C. Cireşan, A. Giusti, L. M. Gambardella, and J. Schmidhuber, "Mitosis detection in breast cancer histology images with deep neural networks," in *Lecture Notes in Computer Science (including subseries Lecture Notes in Artificial Intelligence and Lecture Notes in Bioinformatics)*, 2013, vol. 8150 LNCS, no. PART 2, pp. 411–418.

[28] W. Xie, J. A. Noble, and A. Zisserman, "Microscopy cell counting and detection with fully convolutional regression networks," *Comput. Methods Biomech. Biomed. Eng. Imaging Vis.*, vol. 6, no. 3, pp. 283–292, May 2018.

[29] N. Wahab, A. Khan, and Y. S. Lee, "Two-phase deep convolutional neural network for reducing class skewness in histopathological images based breast cancer detection," *Comput. Biol. Med.*, vol. 85, no. March, pp. 86–97, Jun. 2017.

[30] K. Paeng, S. Hwang, S. Park, and M. Kim, "A unified framework for tumor proliferation score prediction in breast histopathology," in *Lecture Notes in Computer Science (including subseries Lecture Notes in Artificial Intelligence and Lecture Notes in Bioinformatics)*, 2017, vol. 10553 LNCS, pp. 231–239.

[31] C. Li, X. Wang, W. Liu, and L. J. Latecki, "DeepMitosis: Mitosis detection via deep detection, verification and segmentation networks," *Med. Image Anal.*, vol. 45, pp. 121–133, 2018.

[32] C. Li, X. Wang, W. Liu, L. J. Latecki, B. Wang, and J. Huang, "Weakly supervised mitosis detection in breast histopathology images using concentric loss," *Med. Image Anal.*, vol. 53, pp. 165–178, 2019.

[33] M. Macenko *et al.*, "A method for normalizing histology slides for quantitative analysis," in *Proceedings - 2009 IEEE International Symposium on Biomedical Imaging: From Nano to Macro, ISBI 2009*, 2009.

[34] N. Wahab, A. Khan, and Y. S. Lee, "Transfer learning based deep CNN for segmentation and detection of mitoses in breast cancer histopathological images,"

*Microscopy*, vol. 68, no. 3, pp. 216–233, 2019.

[35] K. He, X. Zhang, S. Ren, and J. Sun, "Deep Residual Learning for Image Recognition," *Multimed. Tools Appl.*, vol. 77, no. 9, pp. 10437–10453, Dec. 2015.

[36] T. Y. Lin, P. Dollár, R. Girshick, K. He, B. Hariharan, and S. Belongie, "Feature pyramid networks for object detection," in *Proceedings - 30th IEEE Conference on Computer Vision and Pattern Recognition, CVPR 2017*, 2017.

[37] Qiang Yang, S. J. Pan, Q. Yang, and Q. Y. Fellow, "A Survey on Transfer Learning," *IEEE Trans. Knowl. Data Eng.*, vol. 1, no. 10, pp. 1–15, 2008.

[38] Y. LeCun, K. Kavukcuoglu, C. C. Farabet, and others, "Convolutional networks and applications in vision," in *ISCAS*, 2010, vol. 2010, pp. 253–256.